\documentclass[10pt,twocolumn,letterpaper]{article}

\usepackage{cvpr}      

\definecolor{cvprblue}{rgb}{0.21,0.49,0.74}
\usepackage[pagebackref,breaklinks,colorlinks,allcolors=cvprblue]{hyperref}

\usepackage{booktabs}
\usepackage{multirow}
\usepackage{siunitx}
\usepackage{tcolorbox} 
\usepackage{multicol}  
\usepackage{caption}   
\usepackage{pifont} 
\usepackage[T1]{fontenc}

\usepackage[accsupp]{axessibility} 

\def\paperID{61} 
\def\confName{ICCV}
\def\confYear{2025}
\title{MV-MLM: Bridging Multi-View Mammography and Language for Breast Cancer Diagnosis and Risk Prediction }

\author{
Shunjie-Fabian Zheng\textsuperscript{1,2} \quad
Hyeonjun Lee\textsuperscript{2} \quad
Thijs Kooi\textsuperscript{2} \quad
Ali Diba\textsuperscript{2} \\
\textsuperscript{1}Department of Medicine I, LMU University Hospital, LMU Munich, Germany\\
\textsuperscript{2}Lunit Inc.\\
{\footnotesize\texttt{shunjiefabian.zheng@med.uni-muenchen.de, \{hyeonjun1882, tkooi, ali\}@lunit.io}}
}

\begin{document}
\maketitle
\begin{abstract}
Large annotated datasets are essential for training robust Computer-Aided Diagnosis (CAD) models for breast cancer detection or risk prediction. However, acquiring such datasets with fine-detailed annotation is both costly and time-consuming. Vision-Language Models (VLMs), such as CLIP, which are pre-trained on large image-text pairs, offer a promising solution by enhancing robustness and data efficiency in medical imaging tasks. This paper introduces a novel Multi-View Mammography and Language Model for breast cancer classification and risk prediction, trained on a dataset of paired mammogram images and synthetic radiology reports.
Our MV-MLM leverages multi-view supervision to learn rich representations from extensive radiology data by employing cross-modal self-supervision across image-text pairs. This includes multiple views and the corresponding pseudo-radiology reports. We propose a novel joint visual-textual learning strategy to enhance generalization and accuracy performance over different data types and tasks to distinguish breast tissues or cancer characteristics(calcification, mass) and utilize these patterns to understand mammography images and predict cancer risk.
We evaluated our method on both private and publicly available datasets, demonstrating that the proposed model achieves state-of-the-art performance in three classification tasks: (1) malignancy classification, (2) subtype classification, and (3) image-based cancer risk prediction. Furthermore, the model exhibits strong data efficiency, outperforming existing fully supervised or VLM baselines while trained on synthetic text reports and without the need for actual radiology reports.
\end{abstract}    
\section{Introduction}
\label{sec:intro}
Breast cancer is the most common form of cancer among women in the developed world, with early detection being critical for improving patient outcomes~\cite{siegel2023cancer}. Mammography remains the primary imaging modality for screening for breast cancer, but interpreting mammograms is challenging~\cite{{myers2015benefits}}, and cancers are missed that were visible in hindsight. Computer-aided diagnosis (CAD) systems have been developed to assist radiologists. Still, their performance heavily depends on large-scale annotated datasets, which are expensive and time-consuming to collect. Recent advancements \cite{dembrower2023artificial} have shown promise in automating mammogram analysis. However, state-of-the-art systems still struggle with generalization and data efficiency due to the limited availability of detailed-labeled medical data.

Vision-Language Models (VLMs), such as CLIP \cite{Radford2021LearningTV}, have emerged as a powerful paradigm for learning joint representations of images and text, enabling zero-shot classification, improving data-training efficiency, and providing robust models for different domains. These models have demonstrated success in general computer vision tasks by leveraging large-scale image-text pairs for pre-training. In the medical domain, VLMs have been applied primarily to chest X-rays (CXR), where paired image-report datasets like MIMIC-CXR~\cite{Johnson2019MIMICCXR} are available at scale \cite{zhang2022contrastive}. However, their application to other domains, such as mammography, has been limited due to the high-resolution nature of mammograms and the lack of large-scale paired clinical image-report datasets.

This work proposes a novel Vision-Language Contrastive Learning training model designed for breast cancer classification and image-based risk assessment in mammograms. Our method addresses two key challenges: (1) the scarcity of paired mammogram-report datasets and (2) the need for high-resolution, multi-view image analysis to capture fine-grained visual details critical for accurate diagnosis. To overcome these challenges, we introduce a synthetic report generation approach that leverages tabular metadata from 2D mammography exams (e.g., BI-RADS scores, mass size, calcification type) to create textual descriptions that simulate radiology reports. This allows us to train our model on broader mammographic attributes without relying solely on paired image-report data.

Using contrastive learning, our model builds upon CLIP by aligning high-resolution mammogram images with synthetic text reports in a more rich representation space. This enables our model to learn robust representations that generalize well across multiple downstream tasks, including malignancy, mass and calcification classification as well as breast cancer risk prediction. Furthermore, we demonstrate that our approach outperforms fully supervised and self-supervised learning (SSL) models on these tasks by improving data efficiency and reducing reliance on manual text reports.\\

The main contributions of this paper are as follows:
\begin{itemize}
    \item Multi-View Vision-Language Contrastive Learning Model: We propose a novel VLM training model that aligns high-resolution, multi-view mammogram images with synthetic text reports generated from tabular annotations. This approach enables effective learning from sparsely labeled data without real-world clinical text reports while maintaining high diagnostic accuracy in different downstream tasks and datasets with a generalized model. Our model offers the advantages of using feature map tokenization and Transformer modules with standard ConvNet backbones to maximize the model's efficiency with high-resolution images regarding computation and robustness.
    \item Synthetic Report Generation: We introduce a method for generating synthetic radiology reports based on structured tabular annotations from mammography exams. This allows us to augment existing datasets with textual descriptions that simulate real-world radiology reports to train a more robust vision-language model.
    \item Improved Performance Across Multiple Tasks: Our model achieves state-of-the-art performance on several downstream tasks relevant to breast cancer screening:  malignancy classification, mass and calcification classification, and breast cancer risk prediction. We demonstrate significant improvements over other CLIP-based models, SSL approaches, and fully supervised models.
    \item Data Efficiency and Generalization: By using contrastive learning with synthetic reports, our model demonstrates strong generalization across different datasets. Additionally, experiments show that our approach reduces forgetting during fine-tuning while requiring fewer training parameters and labeled examples compared to traditional supervised methods.\\

Through extensive experiments on publicly available datasets such as VinDr-Mammo~\cite{vindr} and RSNA-Mammo~\cite{rsnamm}, we show that our method improves accuracy and robustness on multiple downstream tasks and is highly generalizable.   
\end{itemize}
 
\section{Related Work}
\label{sec:relatedwork}

\textbf{Vision-Language Models in Medical Imaging:}
Vision Language Models (VLMs), such as CLIP~\cite{Radford2021LearningTV}, which align image and text representations in a joint embedding space, have demonstrated significant benefits in general computer vision tasks, including improved generalizability and reduced reliance on large-scale labeled. The integration of VLMs into medical imaging has shown promise in addressing data efficiency, robustness, and interpretability challenges.

In the domain of medical VLMs for chest X-rays, ConVIRT~\cite{zhang2022contrastive} pioneers the use of contrastive learning to align scans with their corresponding reports.
Building on this studies such as LoVT~\cite{muller2022joint} and GLoRIA~\cite{Huang2021GLoRIA} aim to incorporate global-local representations to enable fine-grained VLMs, enhancing the model's ability to capture detailed features.
On the other hand, MedCLIP~\cite{Wang2022MedCLIP} explores learning vision language models from unpaired medical scans and reports, addressing the scarcity of aligned datasets in medical imaging.
Some works have integrated explicit medical domain knowledge into VLMs; for example, MedKLIP~\cite{wu2023medklip} utilizes structured triplets extracted from reports, while Align~\cite{chen2022align} leverages the Unified Medical Language System (UMLS) to inform and structure training. 
Recently, Kumar et al.~\cite{kumar2024improving} incorporates radiologists' eye-gaze information to reduce the modality gap between image-text pairs, further enriching learned representations. 
CPLIP~\cite{javed2024cplip} and PathAlign~\cite{ahmed2024pathalign} extend VLM applications to histopathology, using comprehensive alignment methods for Whole Slide Images (WSI) and textual descriptions to support interpretability and downstream task in pathology such as image retrieval, WSI classification. 
Our approach mainly focuses on breast imaging data, leveraging pseudo reports generated from metadata instead of actual reports to construct a VLM, considering real-world cases where actual report pairs may not be available.


\textbf{CLIP Model for Mammography:}
To address these challenges, Ghosh et al.\cite{Ghosh2024MammoCLIP} introduced Mammo-CLIP, the first VLM pre-trained specifically on paired mammogram-report data. Mammo-CLIP builds on the CLIP architecture but adapts it for high-resolution mammographic images by employing multi-view supervision (MVS) and data augmentation strategies tailored to the medical domain. The model leverages a screening mammogram dataset paired with real-world radiology reports to enhance generalizability from limited data while maintaining high resolution during training. Additionally, Mammo-CLIP introduces a novel feature attribution method called Mammo-FActOR, which aligns visual features with textual descriptions from radiology reports at a sentence-level granularity. This approach improves interpretability by providing spatially aligned heatmaps that localize important mammographic attributes without relying on ground-truth bounding boxes.

Mammo-CLIP has demonstrated superior performance to baseline models like ResNet-50 and EfficientNet-B5 across various tasks such as classifying mass, calcifications, and breast density. The model's ability to perform zero-shot classification further underscores its robustness in handling out-of-distribution data—a crucial capability for real-world clinical applications where labeled data may be scarce.

\textbf{Breast Cancer Detection \& Risk Prediction:}
In addition to VLMs like Mammo-CLIP, other AI-based methods have been explored for breast cancer detection using mammogram images~\cite{kooi2017large, wu2019deep, liu2020cross, sun2022transformer, rangarajan2022ultra, jain2024follow, Jai_MMBCD_MICCAI2024}.
To effectively capture mammographic features, approaches like multi-scale processing~\cite{rangarajan2022ultra}, utilizing morphological relation between Craniocaudal (CC) and Mediolateral Oblique (MLO) mammogram views~\cite{liu2020cross, jain2024follow} have been employed. 
Moreover, there have been efforts to leverage three-dimensional imaging to further improve breast cancer detection using Digital Breast Tomosynthesis (DBT)~\cite{lee2023transformer, kassis2024detection}. These models employ Vision Transformers (ViTs)~\cite{dosovitskiy2021an} with transfer learning to classify abnomalities across multiple views of DBT scans. While DBT offers enhanced lesion visibility compared to traditional two-dimensional mammography, its widespread adoption is limited due to higher costs and longer acqusition time. 

Rather than classifying mammograms for current signs of breast cancer, image based risk assessment tools \cite{yala2021toward, lee2023enhancing} predict the risk that a patient will develop breast cancer in the future. This risk score can then be used to tailor screening recommendations like a shorter interval or an additional exam. State-of-the art methods for risk prediction from mammograms make use of a hybrid CNN-transformer module with an additive hazard loss for predicting risk at different time points. 
Utilizing our vision encoder results in consistent and significant performance improvements across multiple downstream tasks, such as breast cancer detection and breast cancer risk prediction.

\textbf{Challenges and Promising Directions:}
While VLMs like Mammo-CLIP represent a significant step forward in breast cancer detection through multimodal learning, several challenges remain. First, the availability of large-scale paired datasets for training remains a bottleneck. Although data augmentation techniques can somewhat mitigate this issue, further research is needed to generate synthetic data or leverage weak supervision from unpaired datasets. Additionally, improving model interpretability remains a crucial concern for clinical adoption. Methods like Mammo-FActOR \cite{Ghosh2024MammoCLIP} that provide spatially aligned visual explanations are promising but require further validation across diverse populations and imaging conditions.
\section{Method}
\label{sec:method}

\begin{figure*}[t]
    \begin{center}
    \includegraphics[scale=0.88]
    {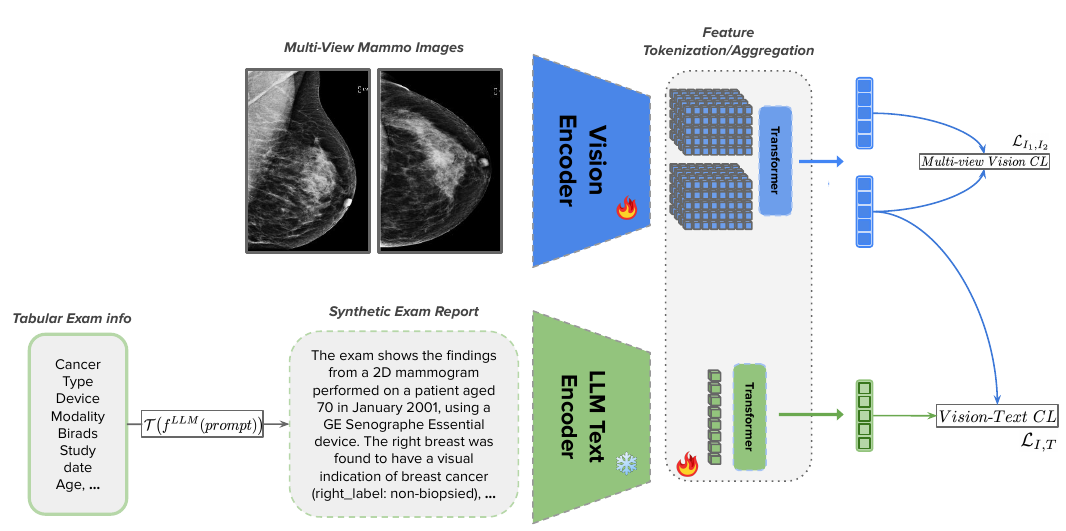}
    \caption{Overview of our proposed \textbf{M}ulti-\textbf{V}iew \textbf{M}ammography-\textbf{L}anguage \textbf{M}odel(MV-MLM) learning for breast cancer screening applications, optimized using objective functions: multi-view visual feature alignment, vision-language contrastive learning by using feature tokenization and aggregation. The model integrates multi-modal inputs, including multi-view mammography exams and synthetic radiology reports, to improve diagnostic and prediction performance in four tasks relevant to breast cancer screening: mass, calcification, malignancy classification, and breast cancer risk prediction.}
    \vspace{-0.6cm}
    \label{fig:method-overview}
    \end{center}
\end{figure*}

A patient's mammography examination consists of four images: two views of the craniocaudal (CC) and the mediolateral oblique (MLO) of each breast, referred to as the laterality. Additionally, the exam contains metadata in tabular form. This metadata has information on a patient and exam level, such as the subject age, gender, and race, is constant for all images, and information about findings at the laterality and view level is specific to a single or pair of images. Then, the set of a patient's examination data is known as exam level data, equipped with $4$ views of the breast tissue and the tabular data, which contains exam level and patient level metadata. A patient-level dataset is a collection of some exam-level data for a patient.

Consider an exam level dataset of size $N$, $\mathcal{D}=\{(x^{I}_{i, lat, view}, x^{tab}_{i})| i\in N, lat\in\{left, right\}, view\in\{MLO, CC\}\}$ consisting of breast mammography exams $x^{I}_{i, lat, view}$ for each view and laterality and tabular annotations $x^{tab}_{i}$. Moreover, the set of tabular data contains patient-level information $x^{tab}_{i, PL}$ specific to each subject and laterality-related information $x^{tab}_{i, lat}$ shared across views for the same $lat$\footnote{For simplicity we will ignore view level annotations, that might very well occur, such as an asymmetry (a finding only visible in one of the two views)}. Therefore we can express the tabular data as $x^{tab}_{i}=(x^{tab}_{i,PL}, x^{tab}_{i, left}, x^{tab}_{i, right})$. Each sample in $\mathcal{D}$ is a set of four tuples $\{(x^{I}_{i,lat,view}, x^{tab}_{i,PL}, x^{tab}_{i, lat})\}$, assuming the clinical findings and annotations are constant across views of the same side of the breast.

\subsection{Pseudo Report Generation}
\label{report_gen}
Inspired by ~\cite{zhao2023investigating}, we first aim to translate the tabular data into synthetic pseudo reports for the mammogram images to enable VLM pre-training.

Let $C$ denote a subset of annotations present in $x^{tab}_{i}$, such that $x^{C}_{i, lat}\in x^{tab}_{i}$, that functions as a filter in order to reduce the total amount of recorded annotations and drop trivial ones. We define $x^{T}_{i, lat}$ as the post-processed text generated by a large language model (LLM) $f^{LLM}(\cdot)$. Using $x^{C}_{i,lat}$, we design the prompt for the LLM as:
\begin{align*}
    prompt_{i,lat,C}=(prefix, \{X^{C}_{i,lat}=x^{C}_{i,lat}\}, suffix)
\end{align*}
Where the prompt $prefix$ is a short general instruction on what we desire the output to be. The prompt $suffix$, on the other hand, summarizes the tabular data, gives a high-level overview, describes some keys in the tabular data, and, most importantly, reinforces the $prefix$ again in more detail. $\{X^{C}_{i,lat}=x^{C}_{i,lat}=\}$ denotes the keys in $C$ and their observed realizations. Although the prompt design contradicts ~\cite{zhao2023investigating}, the desired pseudo reports do not require contextual information besides the $suffix$, as a summary of the keys and values is sufficient for the task.

We utilize the prompt without few-shot examples opposing the proposition by ~\cite{moor2023med, zhao2023investigating}, meaning that we do not provide possible target outputs for the LLM to lean on. This is to reduce the amount of input tokens, considering computational requirements. Furthermore, we hypothesized that simple, noisy text supervision without an exact structured form would work well. The task of generating pseudo reports from tabular data also seems simple enough that it does not call for few-shot examples, even though it might enhance the text output.

Lastly, the raw output of LLMs is not directly usable as the models predict the next token only~\cite{vaswani2017attention}. Therefore, we use a post-processing function $\mathcal{T}^{post}$ that removes possible prompt repetition and cuts off the generated text after the first paragraph that is not the input prompt. Hence we obtain the synthetic pseudo reports as:
\begin{align}
    x^{T}_{i, lat} = \mathcal{T}^{post}\big(f^{LLM}(prompt_{i,lat,C})\big)
\end{align}
Note that in this case, the reports for two views of the same laterality are identical, assuming shared visual cues related to clinical findings across different perspectives of the same breast.

\subsection{Image-Text Contrastive Learning}
With the aid of the synthetic reports we rearrange $\mathcal{D}$ into an image-text dataset $\mathcal{D}^{I,T}=\{(x^{I}_{i, lat, view}, x^{T}_{i,lat})| i\in N, lat\in\{left, right\}, view\in\{MLO, CC\}\}$. Hence, there are $4N$ image-text tuples in $\mathcal{D}^{I,T}$. Since each mammogram has an opposing view in the exam, we define the multi-view dataset $\mathcal{D}^{I,T}_{MV}$ with a cardinality of $2N$.

Operating on $\mathcal{D}^{I,T}_{MV}$, we utilize an image mapping function $f^{I}_{\theta_I}: \mathcal{T}(x^{I}_{i, lat, view}) \to \mathcal{F}_{i,lat,view}^{I}, \mathcal{F}_{i,lat,view}^{I} \in \mathbb{R}^{C\times H\times W}$, projecting an input image to its feature map $\mathcal{F}_{i,lat,view}$, where $C$ is the channel dimension, $H$ and $W$ are the height and width of the feature map, respectively. Note that the size of $H$ and $W$ depend on the input image resolution. $\mathcal{F}_{i,lat,view}$ contains spatial information, indicating where features occur in an image, as well as the local features at each position of the image. Following~\cite{chen2020simple}, we utilize augmented images $\mathcal{T}(x^{I}_{i, lat, view})$ for robustness.

The text mapping function $f^{T}_{\theta_T}: x^{T}_{i, lat} \to H_{i,lat}^{T}, H_{i,lat}^{T} \in \mathbb{R}^{N_{text-token}\times d_{text-token}}$ projects the text sequence $x^{T}_{i, lat}$ onto $N_{text-token}$ each represented by in a vector of $d_{text-token}$ dimensions. Furthermore, the text tokens $H_{i,lat}^{T}$ are equipped with context-aware fine-grained information, contrary to the single global cls token. Both encoders are also parameterized by $\theta_I$ and $\theta_T$.

Next, both the $\mathcal{F}_{i,lat,view}^{I}$ and the text tokens $H_{i,lat}^{T}$ are tokenized to prepare them for the subsequent transformer module. The feature map $\mathcal{F}_{i,lat,view}^{I}$ is first reshaped and transposed from $\mathcal{R}^{C\times H\times W}$ to $\mathcal{R}^{(H\cdot W)\times C}$, sustaining the channel dimensions and reformulating the number of channels into representation dimensions while defining the product of height and width of $\mathcal{F}_{i,lat,view}^{I}$ to pseudo tokens. Since mammography images have high resolution, we further utilize linear projection $g^I_{\theta_I}$ to generate a computationally efficient number of visual tokens $TokRep^{I}_{i,lat,view} = (t^{I}_{1,i,lat,view}, \dots , t^{I}_{N_{intermediate},i, lat,view})$.

It is evident that neither $N_{intermediate}$ and $N_{text-token}$ nor $C$ and $d_{text-token}$ necessarily have the same dimensionality because $N_{text-token}$ is determined by the number of input tokens of the pseudo reports and both $C$ and $d_{text-token}$ are determined by the respective backbone models. Hence, the text tokens also have to be subjected to linear projection $g^T_{\theta_T}$ to match the dimensionality of $TokRep^{I}_{i,lat,view}$, resulting in $TokRep^{T}_{i,lat,view}$. 

Finally, we apply transformers $Tr_{\theta_I}^{I}$ on the image tokens and $Tr_{\theta_I}^{T}$ on the text tokens. Each modality-specific transformer consists of $n_{Tr}$ transformer block with a multi-head self-attention module~\cite{vaswani2017attention} followed by a projection MLP. The self-attention modules have $n_{heads}$ each. Leveraging a global max pooling layer gets the embedding representations $z_{i,lat,view}^{I}\in \mathbb{R}^C$ and $z_{i,lat,view}^{T}\in \mathbb{R}^C $, where the dimensionality is naturally the channel dimension $C$. 

This enables the basic CLIP objective~\cite{Radford2021LearningTV}, aligning image and text embeddings. For simplicity's sake, we consider each tuple of image-text data as its own sample from $2N$. We thus can define the image-text contrastive loss for a mini-batch of size $B$ as the average cross-entropy loss over softmax scaled cosine similarities between image and text representations. 

\begin{align}
\label{clip}
\begin{split}
    \mathcal{L}_{I,T} = &\frac{-1}{2B} \sum_{i=1}^{B} log\bigg\{\frac{exp(z^{I}_{i}(z_{i}^{T})^{'}/\tau_1)}{\sum_{j=1}^{B}exp(z^{I}_{i}(z_{j}^{T})^{'}/\tau_1)}\bigg\}\\
    & \frac{-1}{2B} \sum_{i=1}^{B} log\bigg\{\frac{exp(z^{T}_{i}(z_{i}^{I})^{'}/\tau_1)}{\sum_{j=1}^{B}exp(z^{T}_{i}(z_{j}^{I})^{'}/\tau_1)}\bigg\}
    \end{split}
\end{align}

Where $exp(z^{I}_{i}(z_{i}^{T})^{'})$ is the dot product of normalized vectors projecting onto a hyper-sphere of unit length. $\tau_1$ is the temperature scaling. 

The first term in $\mathcal{L}_{I,T}$ pulls a paired image-text pair together while pushing all over text embeddings within the batch away. Analogously, every image outside the paired sample in the mini-batch is moved away from the text embeddings while pulling its corresponding image embedding close, encouraging a structured order of similar images in the joint embedding space supervised by the pseudo reports. 

\subsection{Multi-View Contrastive Learning}
We are inspired by the alignment of different views of the same laterality, as the MLO and CC views in a mammography exam provide rich visual cues that are both robust and salient across various perspectives. Considering $\mathcal{D}^{I,T}$ we group the two views of each laterality of a patient to a dataset of size $2N$. Then, it is trivial to notice that the multi-view contrastive loss can be defined as:

\begin{align}
\label{multi-view-loss}
    \begin{split}
    \mathcal{L}_{I,I} =& \frac{-1}{2B} \sum_{i=1}^{B} log\bigg\{\frac{exp(z^{I}_{i,MLO}(z_{i, CC}^{I})^{'}/\tau_2)}{\sum_{j=1}^{B}exp(z^{I}_{i, MLO}(z_{j, CC}^{I})^{'}/\tau_2)}\bigg\} \\
    & \frac{-1}{2B} \sum_{i=1}^{B} log\bigg\{\frac{exp(z^{I}_{i, CC}(z_{i, MLO}^{I})^{'}/\tau_2)}{\sum_{j=1}^{B}exp(z^{I}_{i, CC}(z_{j,MLO}^{I})^{'}/\tau_2)}\bigg\}
    \end{split}
\end{align}


Where $\tau_2$ is a temperature scaling again. $\mathcal{L}_{I,I}$ is capable of learning crucial visual attributes that are visible from both views. This objective forces the network to focus on fine-grained constant information between different views. Since the views show various positions of the breast, the model will learn the features present in both views and enhance the handling of noise and visual artifacts that are often present in medical imaging. Therefore, this reinforces the model's generalization abilities.

\subsection{Multi-Task Contrastive Learning}
We define the multi-view CLIP objective on a triplet of two image embeddings and language embeddings as $\text{MV-CLIP}=\mathcal{L}_{I,I} (z^{I}_{i,lat,view_1}, z^{I}_{i,lat,view_2}, \tau_2) + \mathcal{L}_{I,T}(z^{I}_{i,lat,view_1}, z^{T}_{i,lat}, \tau_1)$. As \ref{multi-view-loss} already matches the views and only one text exists per image pair, running only one CLIP loss is sufficient. To introduce variation in the CLIP loss, the views are inter-changed with a probability of $0.5$ during training.

By having no proportions, we ensure that the model simultaneously learns semantic alignment and fine-grained visual consistency with equal contributions.
\section{Experiments}
The experiments section discusses all aspects of data used in the evaluations, implementation, model, state-of-the-art comparison, and ablation studies. 
\label{sec:experiments}

\begin{table*}[!ht]
    \centering
    \scalebox{0.80}{
    \begin{tabular}{l|l|c|c|c|c|c|c|c|c}
    \toprule
    \multirow{2}{*}{Model} &
        \multirow{2}{*}{Encoder} &
        \multicolumn{4}{c|}{Mass} &
        \multicolumn{4}{c}{Calcification}\\
    \cline{3-10}
        &  
        & LP (0.1) & LP (0.5) & LP (1) & FT 
        & LP (0.1) & LP (0.5) & LP (1) & FT\\
    \midrule

    Supervised & RN.34 &
    0.5090 & 0.5796 & 0.5734 & 0.8103 &
    0.4500 & 0.5701 & 0.6377 & 0.9685  \\

    (Custom-)CLIP & RN.34 & 
    0.4765 & 0.5570 & 0.5759 & 0.7952 &
    0.3615 & 0.6848 & 0.7088 & 0.9615 \\

    MV-CLIP & RN.34 & 
    0.5221 & 0.5718 & 0.5868 & 0.8095 &
    0.7820 & 0.6459 & 0.8503 & 0.9637  \\
    \midrule
    Mammo-CLIP~\cite{Ghosh2024MammoCLIP} & EN.B5 &
    0.6040 & 0.6418 & 0.6228 & 0.8312 & 
    0.6399 & 0.6748 & 0.7318 & 0.9746  \\

    Supervised & EN.B5 &
    0.5784 & 0.6384 & 0.6319 & 0.8326 &
    0.6380 & 0.7011 & 0.7075 & 0.9654  \\
    
    \midrule
    (Custom-)CLIP & EN.B5 & 
    0.6797 & 0.7145 & 0.6802 & 0.8231 &
    0.6399 & 0.6772 & 0.8962 & 0.9768 \\
    
    
    MV-CLIP & EN.B5 & 
    0.6941 & \textbf{0.7455} & 0.7536 & 0.8514 &
    0.8887 & 0.9258 & 0.9312 & 0.9787  \\

    (Custom-)CLIP + Tr & EN.B5 & 
    0.6914 & 0.7353 & 0.7562 & 0.8599 &
    0.8402 & 0.8894 & 0.9253 & 0.9803 \\

     MV-CLIP + Tr & EN.B5 & 
    \textbf{0.7083} & 0.7421 & \textbf{0.7649} & \textbf{0.8614} &
    \textbf{0.8558} & \textbf{0.9288} & \textbf{0.9393} & \textbf{0.9812}  \\

    
    

    

    \bottomrule
    \end{tabular}}
    \caption{Classification performance on the VinDr dataset for binary (mass and calcification) with the best performance bolded. The binary classifiers are evaluted with the area under the curve (AUC). We utilize linear probing (LP) with full training set (1) and a semi supervised setting at 10 (0.1) or 50\% (0.5), as well as fine-tuning (FT) for the evaluation. The (Custom-)CLIP model is trained on the same data as MV-CLIP and uses an equally high resolution, while initialized with weights from training on ImageNet.}
    \label{tab:vindr_cls}
\end{table*}
\begin{table}[!ht]
    \centering
    \scalebox{0.8}{
    \begin{tabular}{l|l|c | c | c}
    \toprule
    \multirow{2}{*}{Model} &
        \multirow{2}{*}{Encoder} &
        \multicolumn{3}{c}{Malignancy} \\
        \cline{3-5}
        &  
        & LP (0.1)  & LP (1) & FT\\
    \midrule
        Supervised & RN.34 & 
        0.4949 & 0.5274 & 0.7056  \\

        (Custom-)CLIP & RN.34 & 
        0.5668  & 0.6558 & 0.7423  \\
        MV-CLIP & RN.34 & 
        0.5538  & 0.7400 & 0.7529  \\
        \midrule
        MaMa-CLIP~\cite{du2024multi} & ViT-B-14 & 
        -  & - & 0.73  \\

        MGCA~\cite{wang2022multi} & ViT-B-14 & 
        -  & - & 0.687  \\

        MM-MIL~\cite{wang2023using} & ViT-B-14 & 
        -  & - & 0.65  \\
    
        Mammo-CLIP~\cite{Ghosh2024MammoCLIP} & EN.B5 & 
        0.5411 &  0.6017 & 0.7257 \\

        Supervised & EN.B5 & 
        0.5136 & 0.6077 & 0.7271  \\
        \midrule
        (Custom-)CLIP & EN.B5 & 
        0.5971 & 0.7278 & 0.7659  \\

        MV-CLIP & EN.B5 & 
        0.6714  & 0.7393 & 0.7620  \\


        (Custom-)CLIP + Tr & EN.B5 & 
        0.6383 & 0.7332 & 0.7665  \\

        
        MV-CLIP + Tr & EN.B5 & 
        \textbf{0.6863} & \textbf{0.7406} & \textbf{0.7753} \\

        



       \bottomrule
    \end{tabular}}
    \caption{Malignancy classification performance on the RSNA image level dataset utilizing linear classifier on top of the networks. The AUC is used as the metric. We evaluate the models using linear probing (LP) in a semi-supervised setting, utilizing either 10\% (0.1) of the training set or the entire training set (1).
    The full models are also evaluated with fine-tuning (FT). All the CLIP-based models in the experiments are trained with our data from scratch. (Custom-)CLIP is our CLIP model pre-trained on our data and with a resolution of $(1520,912)$}
    \vspace{-0.5cm}
    \label{tab:rsna_cls}
\end{table}

\subsection{Datasets} We pre-trained our models on a proprietary dataset of 134,500 mammography exams, comprising 540,000 images from four standard views per exam (CC and MLO for both breasts). This large-scale dataset captures a wide range of breast tissue variations and abnormalities, providing a rich foundation for learning complex patterns in mammographic imagery, which enhances the model's generalization ability. As mentioned in the method, this data does not include clinical report text and only has high-level patient and exam information in tabular format.
For the models' evaluation, we have used VinDr-Mammo~\cite{vindr} and RSNA-Mammo~\cite{rsnamm} public datasets for mass, calcification, and malignancy classification tasks and part of our private data for the risk prediction task. The VinDr dataset includes 5,000 exams with 20,000 images from Vietnam, and RSNA-Mammo has 11,913 exams. Our private dataset for risk prediction(risk-mammo) consists of 16,867 exams as the training set and 2245 exams for testing. 

\subsection{Implementation details}

\textbf{Image Transformation:} The grey scale mammograms are loaded as RGB images with 3 color channels. We first turn pixel values $<40$ in the mammograms to zero, as it denotes the background~\cite{Ghosh2024MammoCLIP}. Then, a breast region cropping is applied to isolate the breast before resizing the images to the working resolution of $[1520, 912]$. The breast region cropping consists of edge detection via a classical Sobel filter and a connected component analysis. Following~\cite{chen2020simple,zhang2022contrastive} we further augment the cropped image by affine transformation with rotations up to 20 degrees, a minimum translation of 0.1\%, scaling factors [0.8, 1.2], and shearing by 20 degrees and elastic transformations with ($\alpha = 10, \sigma = 5$), which were proposed by~\cite{Ghosh2024MammoCLIP}. We set $\tau_1=0.007$, $\tau_2=\tau_3=0.1$.

\textbf{Pseudo Report Generation:}
The synthetic pseudo reports are generated by LLaMa-3-7B-instruct. The prefix and suffix of the prompt are displayed in supplementary material, as well as the relevant set of annotations $C$ and sample pseudo reports.

\textbf{Network Architectures:}
For the text encoder, we choose BioClinicalBERT~\cite{alsentzer2019publicly} and freeze it as the representations obtained by the model were empirically found to be sufficient~\cite{moor2023med,Ghosh2024MammoCLIP}. Freezing BioClinicalBERT also reduces the computational burden and since we are mainly interested in the vision model, there is no need to fine-tune it. We utilize different convolutional networks as the image encoder, namely ResNet-34~\cite{he2016deep} and EfficientNet-B5~\cite{Tan2019EfficientNet}. The feature map and the textual tokens are projected onto $256$ tokens. The Transformers consist of $4$ blocks with $8$ self-attention heads each. The MLP within the transformers project onto $1024$ hidden dimensions. All network outputs are normalized.

\textbf{Optimization:}
All models are optimized using AdamW~\cite{loshchilov2017decoupled} with a learning rate of $5e$-$5$ and a weight decay of $1e$-$4$. Additionally, a cosine-annealing scheduler with warm-up for $1$ epoch is used~\cite{loshchilovstochastic}. The training was conducted in a distributed data parallelism~\cite{li2020pytorch} setting with mixed-precision on $8$ H100 GPUs. The pre-training consists of $10$ epochs, where models with a ResNet-34 vision encoder had a per-device mini-batch size of $32$. The CLIP model with EfficientNet-B5 was trained with a per-device mini-batch size of $18$, while all other EfficientNet-B5 models used $8$. The classification was trained with $30$ epochs, utilizing a mini-batch size of $96$ per device for ResNet-34 and $16$ (fine-tuning) or $40$ (linear probing) for EfficientNet-B5. We did model finetuning for the risk prediction task with a batch size of 8 per device training for 20 epochs.

\textbf{Learning Tasks:}
We evaluate our model by comparing its performance in solving downstream tasks to ImageNet-initialized weights and evaluating the effectiveness of its classification performance on data on which the models were not trained. 
We evaluate the backbone on four downstream classification tasks.
\begin{itemize}
\item {\bf Mass classification:} where each view is classified as having an abnormal mass or not.
\item {\bf Calcification classification:} where each view is classified as having calcification.
\item {\bf View-level malignancy classification:} where each view is classified as either positive or negative for breast cancer.
\item {\bf View-level risk assessment:} where each view is classified as either positive or negative for developing breast cancer in 2 or 5 years into the future. 
\end{itemize}
The classification is conducted on a frozen vision backbone (linear probing) with both complete training data and smaller portions of it. Then, fine-tuning experiments are conducted to further evaluate the VLM pre-training effectiveness.

\textbf{Baseline Comparison:}
For a fair comparison, several baselines are built. A fully supervised model is trained on our private data with a cancer label for each image. The pre-training is conducted with a binary classifier and weighted cross-entropy loss. Additionally, the pre-trained EfficientNet-B5 backbone from Mammo-CLIP is directly used to solve the aforementioned tasks. It has to be stated that the best-performing Mammo-CLIP model was also trained with one evaluation dataset and actual clinical reports. We also trained a (Custom-)CLIP model on our data in the same manner as described earlier. The (Custom-)CLIP model uses the exact resolution and vision-text dataset as our method. It is not the Open-CLIP model with their weights, as the low resolution is not suitable for mammography images\cite{Ghosh2024MammoCLIP}. Lastly, to fully explore the effectiveness of our process, we run both the (Custom-)CLIP and MV-CLIP settings without the transformers and directly extract embeddings from the vision encoder for the contrastive objectives. EfficientNet embeddings are obtained by pooling the feature map.

\subsection{Results}
The classification performances on different tasks for our multi-view contrastive learning methods are presented in Tables \ref{tab:vindr_cls}, \ref{tab:rsna_cls} and \ref{tab:risk}. Moreover, table \ref{tab:ssl} compared our method with Open-CLIP and self-supervised learning (SSL) algorithms.  

\textbf{Breast Mass Classification:} The binary mass classification shows the effectiveness of integrating multiple views during VLM pre-training, as each proposed model surpasses Mammo-Clip, (Custom-)CLIP, and the Supervised baselines for fine-tuned classification tasks. A view on linear probing further reinforces the robustness of our models since we achieved excellent performance even with the data-scarce regime. The linear probe with the full for MV-CLIP further displays the generalization and robustness of learning from synthetic reports and multiple views since it almost rivals the fine-tuned version, supporting the actual applicability of our methods in clinical applications where data is scarce and fine-tuning expensive.

Integrating the transformer on top of the CNN and its tokenized feature map improves the performance even further, while the (Custom-)CLIP setting benefits $3.7$ \% points gained with finetuning. The multi-view framework gains fewer improvements compared to (Custom-)CLIP, indicating that integrating multiple mammography views during pre-training aids in finding generalizable and more optimal representations. Linear probing results could also be improved, with the only exception of LP at $50$ \% data, in which case the MV-CLIP with and without transformer perform comparably. The overall best improvements can be seen during linear probing, suggesting the generalization strength of our models, especially compared to Mammo-CLIP, which utilized the VinDr Mammo dataset during training for the best-performing model. We could show almost $10$\% improvements at $10$ and $50$ \% of the data while over $14$ \% gains with the whole dataset.

\textbf{Breast Calcification Classification:} Calcification classification in Table.\ref{tab:vindr_cls} shows a similar picture to the mass classification. Multi-view settings improve linear probing at any level of training data compared to the baselines, which supports the usefulness of multi-view settings. Although our models reach the best performance for fine-tuned classification, the results suggest saturation in the dataset or that calcification is not too difficult to solve during fine-tuning. Our models can learn robust and generalizable visual information from different views and text. The results from the Mammo-CLIP model in Tables.\ref{tab:vindr_cls} and \ref{tab:rsna_cls} are obtained by using the released model within our evaluation pipeline. 

\begin{table}[!ht]
    \centering
    \scalebox{0.8}{
    \begin{tabular}{l|l|c | c}
    \toprule
    \multirow{2}{*}{Model} &
        \multirow{2}{*}{Encoder} &
        \multicolumn{2}{c}{Malignancy} \\
        \cline{3-4}
        &  
          & LP (1) & FT\\
    \midrule
        SimClr~\cite{chen2020simple} & RN.34 & 
        0.669 &  0.908  \\

        SwaV~\cite{caron2020unsupervised} & RN.34 & 
        0.671 &  0.907  \\

        DINO~\cite{caron2021emerging} & RN.34 & 
        0.665 &  0.909  \\

        BYOL~\cite{grill2020bootstrap} & RN.34 & 
        0.659 &  0.909  \\

        Open-CLIP~\cite{Radford2021LearningTV} & RN.50 & 
        -- &  0.915 \\ \midrule

        (Custom-)CLIP & RN.34 & 
        0.826 &  0.937  \\

        MV-CLIP (ours) & RN.34 & 
        \textbf{0.845} &  \textbf{0.939}  \\

       \bottomrule
    \end{tabular}}
    \caption{Comparison of our method with self-supervised learning methods and the pre-trained Open-CLIP model from OpenAI. The (Custom-)CLIP model is pre-trained on our data using the same resolution as MV-CLIP. We evaluate the malignancy classification performance via AUC after fine-tuning on our private test data.}
    \vspace{-0.5cm}
    \label{tab:ssl}
\end{table}


\textbf{Malignancy Classification in Breasts:} Table.\ref{tab:rsna_cls} contains the malignancy classification performance in which the multi-view objectives reinforce the previous findings. The generalization ability of multi-view contrastive learning for VLMs is outstanding, as the AUC in a scarce setting with a frozen encoder could be improved drastically compared to an isolated (Custom-)CLIP setting and the supervised models, implying the extraction of robust, salient features from noisy text and two views. Fine-tuning the pre-trained models strengthens the benefit of multi-view VLM pre-training as we reach state-of-the-art performances even in comparison to DINO-based transformer models such as MaMa~\cite{du2024multi} and MGCA~\cite{wang2022multi}. Again, the added benefit of the transformer is evident, as it further elevates the multi-view framework. Thereby showing the effectiveness of each module in our method.

Table \ref{tab:ssl} compares malignancy classification performance between our proposed method and several state-of-the-art SSL approaches, including popular contrastive learning and knowledge distillation-based image-only methods and the Open-CLIP pre-trained weights. Our method consistently achieves superior results on both linear probing and finetuning. Specifically, we observe notable improvements in accuracy metrics, demonstrating the robustness and effectiveness of our approach in capturing discriminative features relevant to malignancy detection. These results underline the potential of our proposed method as a strong baseline for future research in self-supervised learning for medical image classification tasks.

\begin{table}[!ht]
    \centering
    \scalebox{0.77}{ 
    \begin{tabular}{l|l|r|r|r}
      \toprule
      Model & Pre-trained Model & C-Index & 2-year AUC & 5-year AUC \\ \midrule
      RN-34 & supervised & 0.65 & 0.69 & 0.64 \\ 
      Mirai~\cite{yala2021toward} & supervised & 0.57 & 0.62 & - \\ 
      RN+Tr & supervised & 0.69 & 0.74 & 0.65 \\ \midrule
      RN+Tr & (Custom-)CLIP & 0.71 & 0.73 & 0.64 \\ 
      \textbf{RN+Tr} & \textbf{MV-CLIP (ours)} & \textbf{0.73} & \textbf{0.76} & \textbf{0.69} \\ \bottomrule
    \end{tabular}}
    \caption{Risk prediction performance on our internal risk-mammo dataset. The performance measures are c-index and 2-year and 5-year AUC as standard breast risk prediction model measures. Comparisons are reported based on baselines and the method trained on the VLM pre-trained models. RN+Tr denotes a simpler version of the Mirai model using RN-34 as the encoder and Transformer for the feature aggregation module.}
    \label{tab:risk}
\end{table}
\textbf{Breast Cancer Risk Prediction:} Table \ref{tab:risk} presents the C-Index~\cite{yala2021toward}, 2-year, and 5-year AUC scores for breast cancer risk prediction evaluation on our risk-mammo internal data. The C-index represents the probability that, for a randomly selected pair of patients, the patient who develops breast cancer earlier is assigned a higher risk score by the model than the one who does not.

We showed that our VLM pre-trained methods can outperform similar methods using supervised pre-trained models. By supervised pre-trained models, we mean backbones previously trained on the same amount of data for the malignancy classification task. This indicates that the models were able to learn visual clues related to other risk factors from synthetic text, such as age and breast density, which can improve risk prediction performance.

\subsection{Ablation Study}
\begin{table}[!ht]
    \centering
    \scalebox{0.52}{
    \begin{tabular}{l|c | c | c | c | c | c | c | c | c}
    \toprule
    \multirow{3}{*}{Encoder} & \multicolumn{3}{|c}{Private Data 1} &  \multicolumn{3}{|c}{Private Data 2} & \multicolumn{3}{|c}{Private Data 3} \\
    \cline{2-10}
     & \multicolumn{3}{|c}{Models} & \multicolumn{3}{|c}{Models} & \multicolumn{3}{|c}{Models} \\
    \cline{2-10}
         &
        CLIP* &
        SigLIP &
        MV-CLIP &
        CLIP* &
        SigLIP &
        MV-CLIP &
        CLIP* &
        SigLIP &
        MV-CLIP\\
    \midrule
    RN.34 & 0.9492 & 0.8569 & \textbf{0.9694} & 0.8661 & 0.8078 & \textbf{0.9200} & 0.9257 & 0.8304 & 0.9536 \\

    RN.50 & 0.9465 & 0.8546 & 0.9656 & 0.8627 & 0.8204 & 0.9164 & 0.9188 & 0.8206 & 0.9521 \\

    EN.B2 & 0.9419 & 0.8708 & 0.9575 & 0.8477 & 0.8266 & 0.8899 & 0.9158 & 0.8607 & 0.9438 \\

    EN.B3 & 0.9519 & \textbf{0.8816} & -- & 0.8813 & \textbf{0.8544} & -- & 0.9324 & \textbf{0.8738} & --\\

    EN.B5 & \textbf{0.9581} & -- & 0.9628 & \textbf{0.8879} & -- & 0.9134 & \textbf{0.9447} & -- & \textbf{0.9560}\\
       \bottomrule
    \end{tabular}}
    \caption{Zero Shot image retrieval performance of ResNets and EfficientNets for our datasets. We report the Recall at 1. We compare (Custom-)CLIP (denoted by CLIP*) and MV-CLIP models to SigLIP~\cite{zhai2023sigmoid}.}
    \label{tab:nets-n-losses}
\end{table}

We have evaluated and studied the effects of different backbone architectures and also variations of objective functions, such as sigmoid-based contrastive loss. Table.\ref{tab:nets-n-losses} displays the zero-shot retrieval performance of different vision backbones and driving objectives. As it is evident that ResNet-50 variants perform worse than their smaller counterparts, we focus on ResNet-34 for this class of convolutional networks. Moreover, the same argument leads us to mainly report EfficientNet-B5-based model results in the previous comparisons as the best setup. One of the other findings in our study is that Sigmoidal loss functions~\cite{zhai2023sigmoid} perform significantly worse compared to regular (Custom-)CLIP settings for the data at hand. This suggests that sigmoid-based loss in VLM setup is insufficient to handle fine-grained details of medical images compared to softmax loss and can not be generalized well for such domain-specific problems.\\


\section{Conclusion}


In this paper, we introduced MV-MLM, a Multi-View Vision-Language Contrastive Learning model designed for breast cancer/anomaly detection and risk prediction from mammography images. Our method addresses the limited availability of paired mammogram-report datasets by aligning high-resolution mammograms with synthetic text reports generated from structured annotations. MV-MLM outperformed existing CLIP-based, SSL, and fully supervised methods on malignancy classification, breast mass and calcification estimation, as well as risk prediction tasks across public datasets (VinDr-Mammo and RSNA-Mammo). The strong performance and generalization capabilities demonstrate MV-MLM's potential for clinical applications, especially considering the the transferability shown during linear probing. Future work includes extending our approach to other imaging modalities and enhancing interpretability for modalities with limited annotated data.
{
    \small
    \bibliographystyle{ieeenat_fullname}
    \bibliography{main}
}


\end{document}